\documentclass[journal]{IEEEtran}
\usepackage{amsmath,amsfonts}
\usepackage[ruled,linesnumbered]{algorithm2e}
\usepackage{array}
\pdfoutput=1 

\ifCLASSOPTIONcompsoc
\usepackage[caption=false, font=normalsize, labelfont=sf, textfont=sf]{subfig}
\else
\usepackage[caption=false, font=footnotesize]{subfig}
\fi
\usepackage{textcomp}
\usepackage{stfloats}
\usepackage{url}
\usepackage{verbatim}
\usepackage{graphicx}
\usepackage{cite}
\usepackage{amssymb}
\usepackage{booktabs}
\usepackage{algpseudocode}
\usepackage{multirow}
\usepackage{tabularx}
\usepackage{enumerate}
\usepackage{bigstrut}
\usepackage{dsfont}
\usepackage{float}
\usepackage{makecell}
\usepackage{ragged2e}
\usepackage{mathrsfs}
\usepackage[hidelinks,colorlinks,allcolors=black,pdfstartview=Fit]{hyperref}
\usepackage[capitalize]{cleveref}
\crefname{section}{Sec.}{Secs.}
\Crefname{section}{Section}{Sections}
\Crefname{table}{Table}{Tables}
\crefname{table}{Tab.}{Tabs.}
\newcolumntype{Z}{>{\centering\let\newline\\\arraybackslash\hspace{0pt}}X}
\begin{document}
\title{SISP: A Benchmark Dataset for Fine-grained Ship Instance Segmentation in Panchromatic Satellite Images}
\author{Pengming Feng, \IEEEmembership{Member, IEEE}, Mingjie Xie, Hongning Liu, Xuanjia Zhao, Guangjun He, Xueliang Zhang, \IEEEmembership{Member, IEEE}, Jian Guan, \IEEEmembership{Member, IEEE}
   \thanks{This work was sponsored by Beijing Nova Program (20230484261). (Corresponding author: Jian Guan.)}
  \thanks{Pengming Feng and Guangjun He are with the State Key Laboratory of Space-Ground Integrated Information Technology, CAST, Beijing 100095, China (e-mail: p.feng.cn@outlook.com; hgjun\_2006@163.com).}
  \thanks{Mingjie Xie, Xuanjia Zhao and Jian Guan are with the Group of Intelligent Signal Processing (GISP), College of Computer Science and Technology, Harbin Engineering University, Harbin 150001, China (e-mail: xiemingjie@hrbeu.edu.cn; xuanjia@hrbeu.edu.cn; j.guan@hrbeu.edu.cn).}
  \thanks{Hongning Liu is with the College of Electrical and Mechanical Engineering, Dalian Minzu University, Dalian 116600, China (e-mail: lhn873949437@163.com).}
  \thanks{Xueliang Zhang is with the School of Geography and Ocean Science, Nanjing University, Nanjing 210023, China (email: zxl@nju.edu.cn).}
}
\maketitle
\begin{abstract}
Fine-grained ship instance segmentation in satellite images holds considerable significance for monitoring maritime activities at sea.
However, existing datasets often suffer from the scarcity of fine-grained information or pixel-wise localization annotations, as well as the insufficient image diversity and variations, thus limiting the research of this task.
To this end, we propose a benchmark dataset for fine-grained Ship Instance Segmentation in Panchromatic satellite images, namely SISP, which contains 56,693 well-annotated ship instances with four fine-grained categories across 10,000 sliced images, and all the images are collected from SuperView-1 satellite with the resolution of 0.5m. 
Targets in the proposed SISP dataset have characteristics that are consistent with real satellite scenes, such as high class imbalance, various scenes, large variations in target densities and scales, and high inter-class similarity and intra-class diversity, all of which make the SISP dataset more suitable for real-world applications. 
In addition, we introduce a Dynamic Feature Refinement-assist Instance segmentation network, namely DFRInst, as the benchmark method for ship instance segmentation in satellite images, which can fortify the explicit representation of crucial features, thus improving the performance of ship instance segmentation.
Experiments and analysis are performed on the proposed SISP dataset to evaluate the benchmark method and several state-of-the-art methods to establish baselines for facilitating future research. 
The proposed dataset and source codes will be available at: \url{https://github.com/Justlovesmile/SISP}.
\end{abstract}
\section{Introduction}
\label{sec:intro}
\IEEEPARstart{F}{ine-grained} ship instance segmentation is a fundamental but challenging task in satellite image interpretation, which aims to achieve pixel-wise localization for each ship instance and differentiate its category at a more detailed and specific level \cite{huang2021orientated, sun2022global}. It has shown great potential for various real-world applications, such as maritime surveillance, marine traffic monitoring, port management and disaster response \cite{wei2020hrsid, zhang2022mask, guo2023fine, han2021fine}.

Benefiting from the powerful feature extraction ability of convolutional neural network (CNN) and the publicly available large-scale datasets, e.g., Cityscapes \cite{cordts2016cityscapes} and COCO \cite{lin2014microsoft}, deep learning based instance segmentation methods have made significant improvements in natural images over the past few years \cite{liu2018path, bolya2019yolact, wang2020solov2, xie2020polarmask, cheng2022pointly}.
Driven by the success in natural images, multiple datasets for Earth observation have been proposed \cite{xia2018dota,waqas2019isaid,wei2020hrsid}, and numerous instance segmentation algorithms for satellite images have been developed by fine-tuning models on these datasets \cite{su2020hq,xu2021improved,su2019object}. 
However, ship instance segmentation in satellite images is still facing challenges, due to the demand of simultaneously achieving fine-grained recognition and pixel-wise localization is growing rapidly for practical applications, but existing datasets have the deficiencies for providing annotations that satisfy the demand as well as reflecting the characteristics of real satellite scenes, thereby limiting the research of this task.

Specifically, deep learning is the data-driven technique \cite{sun2022fair1m}, thus sufficient data is essential for specific tasks. In this case, the well-annotated datasets are required for the study of fine-grained ship instance segmentation in satellite images. However, most existing instance segmentation datasets, such as iSAID \cite{waqas2019isaid}, Airbus Ship \cite{airbus-ship-detection} and HRSID \cite{wei2020hrsid}, are poorly annotated with fine-grained categories, which hinders models from learning to better distinguish ships within subordinate categories. Conversely, existing datasets with fine-grained categories, such as ShipRSImageNet \cite{zhang2021shiprsimagenet} and FAIR1M \cite{sun2022fair1m}, are usually designed for object detection, which do not provide pixel-wise localization annotations for targets. Therefore, a relatively specialty dataset, which provides both fine-grained information and accurate pixel-wise annotations is in great demand.

In addition, the performance of deep learning based ship instance segmentation models in real-world applications largely depends on whether the dataset reflects the characteristics of real satellite scenes. However, most of the commonly used datasets for Earth observation are not composed entirely of satellite images, e.g., iSAID \cite{waqas2019isaid} mixes numerous aerial images from Google Earth, which limits the research of intelligent satellite image interpretation for real-world applications. Therefore, a dataset that can reflect the real characteristics of targets and images in practical satellite applications is highly desirable.

\begin{figure*}
    \centering
    \includegraphics[width=0.95\linewidth]{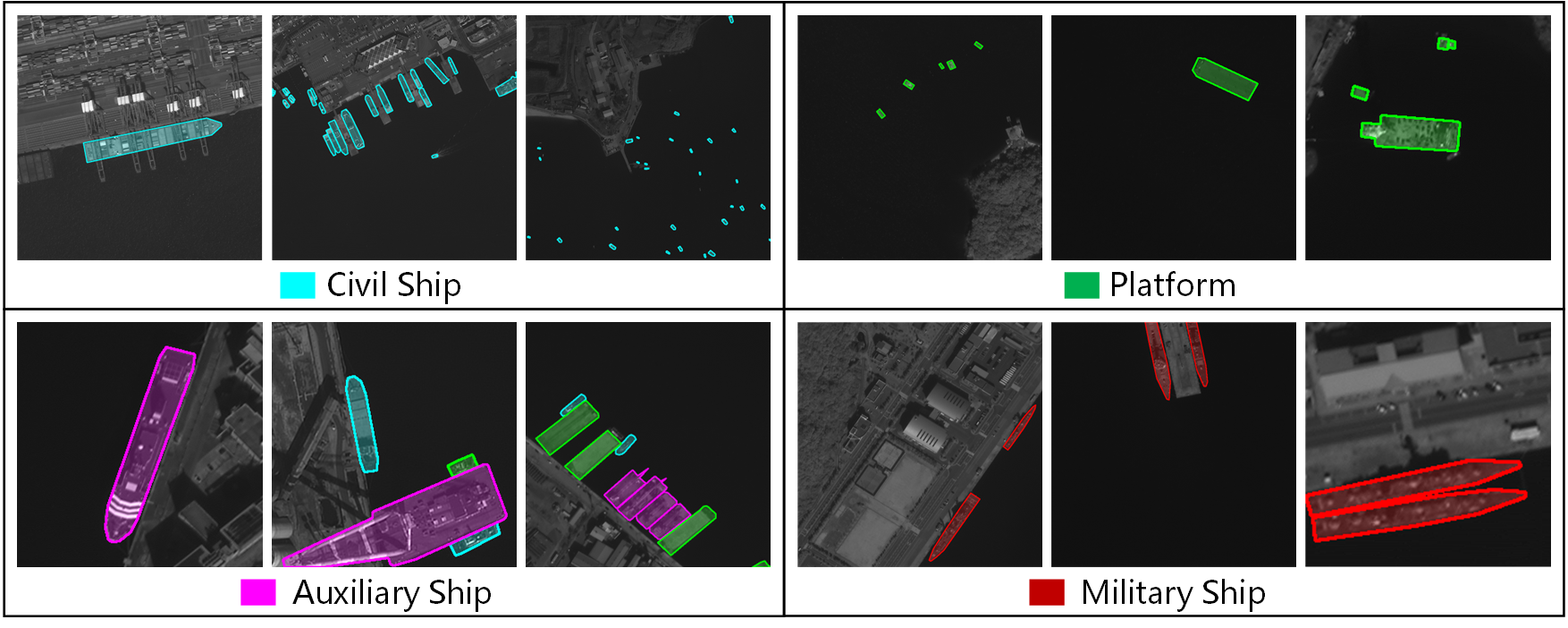}
    \caption{Illustration of typical examples taken from the proposed SISP dataset, where the pixel-wise localization annotations for each category are displayed in the specific colour for better visualization.}
    \label{fig:1_categories}
\end{figure*}

Consequently, to tackle above problems, we have constructed a benchmark dataset for fine-grained Ship Instance Segmentation in Panchromatic satellite images, namely SISP. Compared with existing datasets, our proposed SISP dataset has distinctive properties, which can better benefit the study for fine-grained ship instance segmentation.
Firstly, the SISP dataset is entirely composed of panchromatic satellite images, which contains 56,693 well-annotated ship instances across 10,000 sliced images. All the panchromatic images are collected from SuperView-1 satellite with the resolution of 0.5m, while panchromatic images have a higher spatial resolution compared with other satellite images \cite{xue2017automatic, zhang2016fusion}, thus it is particularly useful for the applications that require detailed feature extraction and analysis. 
Secondly, different from existing ship instance segmentation datasets in satellite images with only one category, our SISP dataset can provide fine-grained categories for ships, i.e., civil ship, auxiliary ship, platform and military ship, as illustrated in Fig.~\ref{fig:1_categories}.
Thirdly, the SISP dataset is more suitable for real-world applications, where targets have characteristics that are consistent with real satellite scenes, such as high class imbalance, various scenes, large variations in target densities and scales, as well as the high inter-class similarity and intra-class diversity.
In addition, we introduce a Dynamic Feature Refinement-assist Instance segmentation network, namely DFRInst, as benchmark method, where a dynamic feature refinement (DFR) module is proposed in our method, which can fortify the explicit representation of crucial features, thus improving the performance for ship instance segmentation in satellite images.
Finally, we evaluate the proposed method and several state-of-the-art instance segmentation algorithms on the SISP dataset to establish baselines for facilitating future research.

The main contributions of this paper are summarized as follows:
\begin{enumerate}
\item A publicly available dataset, namely SISP, is proposed for ship instance segmentation in satellite images. It provides fine-grained ship categories with the characteristics that are consistent with real satellite scenes, which is more suitable for real-world applications.
\item A benchmark method, namely DFRInst, is developed for ship instance segmentation in satellite images, where a DFR module is proposed to fortify the explicit representation of crucial features, thus improving the performance of ship instance segmentation.
\item We evaluate the instance segmentation task with the proposed method and several state-of-the-art algorithms on the SISP dataset to build the baselines for facilitating future research.
\end{enumerate}

The remainder of this paper is organized as follows: Section \ref{sec:related} briefly reviews the related work; Section \ref{sec:dataset} presents the proposed SISP dataset in detail; Section \ref{sec:ourmethod} introduces the proposed benchmark method for ship instance segmentation in satellite images; Section \ref{sec:experiments} gives the benchmark experiments and analysis; Section \ref{sec:conclusion} concludes the paper.

\section{Related Work}
\label{sec:related}

\begin{table*}[t]
\caption{Comparison with other commonly used datasets for Earth observation in satellite images. Here, OBB and HBB represent the oriented bounding box and horizontal bounding box, respectively.}
\resizebox{\linewidth}{!}{
\begin{tabular}{llccccccc}
\toprule
Datasets & Source & Annotations & Instances & Images & \makecell{Image\\width} & Categories & \makecell{All\\satellite} & \makecell{Fine-\\grained} \\
\midrule
NWPU-VHR10 \cite{cheng2014multi}    & Google Earth & OBB    & 3,775   & 800    & $\sim$1,000  & 10 & N & N \\
VEDAI \cite{razakarivony2016vehicle}& Utah AGRC    & OBB    & 3,640   & 1,210  & 512;1,024    & 9  & Y & Y \\ 
UCAS-AOD \cite{zhu2015orientation}  & Google Earth & OBB    & 6,029   & 910    & $\sim$1,000  & 2  & N & N \\
HRSC2016 \cite{liu2017high}         & Google Earth & OBB    & 2,976   & 1,061  & $\sim$1,100  & 22 & N & Y \\
RSOD \cite{long2017accurate} & Google Earth, Tianditu & HBB & 6,950   & 976    & $\sim$1,000  & 4  & N & N \\
DOTA \cite{xia2018dota} & Google Earth, JL-1, GF-2 & OBB    & 188,282 & 2,806  & 800$-$13,000 & 15 & N & N \\
AIR-SARShip-1.0 \cite{xian2019air}  & GF-3         & HBB    & 3,000   & 31     & 3,000        & 1  & Y & N \\ 
DIOR \cite{li2020object}            & Google Earth & HBB    & 192,472 & 23,463 & 800          & 20 & N & N \\
SIMD \cite{haroon2020multisized}    & Google Earth & HBB    & 45,096  & 5,000  & 1024$\times$768 & 15 & N & Y \\
FAIR1M \cite{sun2022fair1m}     & Google Earth, GF & OBB  & 1,020,579 & 42,796 & 600$-$10,000 & 37 & N & Y \\
\midrule
iSAID \cite{waqas2019isaid}      & Google Earth, JL-1, GF-2 & Polygon & 655,451 & 2,806  & 800$-$13,000 & 15  & N & N \\
SpaceNet MVOI \cite{weir2019spacenet} & WorldView-2         & Polygon & 126,747 & 62,000 & 900          & 1   & Y & N \\
HRSID \cite{wei2020hrsid} & Sentinel-1B, TerraSAR-X, TanDEM & Polygon & 16,951  & 5,604  & 800          & 1   & Y & N \\
RarePlanes \cite{shermeyer2021rareplanes} & WorldView-3, Synthetic & Polygon & 644,258 & 50,253 & 1,080 & 110 & N & Y \\
\midrule
SISP (Ours) & SuperView-1 & Polygon & 56,693 & 10,000 & 800 & 4 & Y & Y\\ 
\bottomrule
\end{tabular}}
\label{tab:datasets}
\end{table*}

\subsection{Datasets for Instance Segmentation in Satellite Images}

Instance segmentation, which can be considered as a combination of object detection and semantic segmentation, holds a crucial role in satellite image interpretation \cite{yeh2020using, garnot2021panoptic, neupane2021deep, zorzi2022polyworld}. However, despite numerous publicly available datasets in satellite images have been introduced in the past decade \cite{cheng2014multi, razakarivony2016vehicle, zhu2015orientation, liu2017high, long2017accurate, xia2018dota, xian2019air, li2020object}, as shown in Table \ref{tab:datasets}, few instance segmentation datasets in satellite images are proposed due to the difficulty in pixel-wise annotation. 

The first large-scale benchmark dataset for instance segmentation in aerial images is the iSAID dataset \cite{waqas2019isaid}. It is re-annotated from the DOTA dataset \cite{xia2018dota}, which contains 2,806 images and 655,451 instances with 15 different geospatial object categories. Although this dataset is widely used in instance segmentation research, images in the iSAID dataset are collected from different platforms, e.g., Gaofen-2 (GF-2) satellite, Jilin-1 (JL-1) satellite and Google Earth, which does not consist exclusively of satellite images, hence it is sub-optimal for instance segmentation application in satellite images.
Whereas the SpaceNet MVOI dataset \cite{weir2019spacenet} is composed entirely of satellite images from WorldView-2 satellite, which includes 62,000 images with 126,747 building footprints. Although this dataset contains abundant instances as well as multiple views, it contains only one category, whereas existing studies have shown that targets of interest in satellite images are usually belong to multiple fine-grained categories \cite{sun2022fair1m}.
Furthermore, the HRSID dataset \cite{wei2020hrsid} contains 5,604 cropped synthetic aperture radar (SAR) images with 16,951 instances for ship detection and segmentation. As the first SAR ship dataset which supports instance segmentation, it provides high-resolution SAR images, however, it also has only a single category, i.e., ship.
In addition, the RarePlanes \cite{shermeyer2021rareplanes} is a large-scale dataset for fine-grained plane instance segmentation, which provides 5 features, 10 attributes and 33 sub-attributes for aircraft. Although this dataset contains 644,258 images, most of the images are synthetically generated, which cannot adequately reflect the complexity of the problem in the real world.

Consequently, existing remote sensing datasets for instance segmentation either do not consist exclusively of satellite images or contain only a single category, which are not suitable for real-world applications in complicated satellite scenes. Therefore, considering the significance of precise localization as well as the fine-grained information for instance segmentation in satellite images, we introduce a novel dataset, namely SISP, which is a well-annotated and challenging dataset for fine-grained ship instance segmentation in panchromatic satellite images.

\subsection{Datasets for Fine-grained Recognition in Satellite Images}

Over the past decades, researchers have found that fine-grained recognition in satellite images is a particularly challenging task, due to the large intra-class differences and small inter-class differences \cite{sun2022fair1m, di2021public}.
Therefore, some related datasets with fine-grained information have been proposed. 
For instance, the VEDAI dataset \cite{razakarivony2016vehicle} is a fine-grained vehicle detection dataset with 3,640 instances and 1,210 images, which provides fine-grained information of car categories, e.g., vans, truck and pickup. 
The HRSC2016 dataset \cite{liu2017high} is one of the most commonly used dataset for ship detection in remote sensing images with oriented bounding box (OBB) annotations, which contains 22 classes of ships with only 1,061 images. 
In addition, the SIMD dataset \cite{haroon2020multisized} is a fine-grained multi-class object detection and recognition dataset, which contains seven types of cars, six types of aircraft, boats and other classes. It collects 5,000 images from Google Earth, which contains 45,096 instances. However, it is annotated with horizontal bounding box (HBB), which cannot reflect the orientation information of targets in remote sensing images. 
The FAIR1M dataset \cite{sun2022fair1m} is a large-scale fine-grained object detection dataset in remote sensing images, which includes 37 categories with 42,796 images. It contains more than 1.02 million instances with OBB annotations.

Although the aforementioned datasets can provide fine-grained information and greatly facilitate the development of fine-grained object detection and recognition in satellite images, they do not provide pixel-wise segmentation masks, which limits their applications for tasks that require precise localization. To this end, the proposed SISP dataset applies the pixel-wise polygon annotation and providing the fine-grained information for ship instance segmentation in panchromatic satellite images.

\subsection{Deep Learning based Instance Segmentation Methods}
\label{sec:related_ins}

With the significant breakthroughs in deep learning techniques, numerous CNN based instance segmentation methods have been proposed, which can be divided mainly into two categories according to the network structure, i.e., two-stage and one stage methods. In addition, the Transformer \cite{vaswani2017attention} based methods, involving self-attention mechanism, are showing great advantages on the instance segmentation task. Therefore, attention-based instance segmentation methods have also turned into a popular research direction recently.

\subsubsection{Two-stage Instance Segmentation Methods}
With the distinct sequences in target localization and mask generation, two-stage instance segmentation methods have achieved remarkable success, which can be further divided into top-down (i.e., detection-based) and bottom-up (i.e., segmentation-based) methods \cite{gu2022review}.

The top-down methods adopt the idea of detection first and then segmentation.
As the preliminary study of instance segmentation, SDS \cite{hariharan2014simultaneous} achieves segmentation by classifying the generated mask proposals \cite{girshick2014rich}, which lays the foundation for the subsequent works.
Afterwards, Mask R-CNN \cite{he2017mask} is proposed based on Faster R-CNN \cite{girshick2015fast} but adds a branch for predicting the instance masks, making it a simple but effective method for instance segmentation.
Similar as Mask R-CNN, CenterMask \cite{lee2020centermask} introduces a spatial attention-guided mask branch based on the FCOS \cite{Tian_2019_ICCV}. Although it is built upon the one-stage detector, it still follows the paradigm of detection before segmentation.
Considering image segmentation as a rendering problem, PointRend \cite{kirillov2020pointrend} computes masks in a coarse-to-fine fashion by making predictions over a set of selected points.
However, top-down methods rely on the object detection performance greatly, therefore, researchers have tried to regard the instance segmentation as an image clustering task, i.e., bottom-up methods. 

The main idea of bottom-up methods is to perform pixel-wise semantic segmentation first, and then map each pixel to the corresponding instance.
For instance, SGN \cite{liu2017sgn} employs a series of CNN layers, each of which solves a sub-grouping problem, to compose objects out of pixels.
SSAP \cite{gao2019ssap} computes the probability that two pixels belong to the same instance in a hierarchical fashion based on a pixel-pair affinity pyramid, and then it generates the instances sequentially through the cascade graph partition. Although the bottom-up methods are more intuitive, they heavily rely on the robust semantic segmentation network and require more generalized post-processing methods to distinguish different instances.

In addition, some two-stage methods often employ the cascade structure, such as Cascade R-CNN \cite{cai2019cascade} and HTC \cite{chen2019hybrid}, to further improve the instance segmentation accuracy by fusing more useful information, but require more computing resource consumption.

\subsubsection{One-stage Instance Segmentation Methods}
For more efficiency, researchers have tried to use a more simple network to achieve real-time instance segmentation.
For example, to improve the speed of instance segmentation, YOLACT \cite{bolya2019yolact} divides the whole task into two parts. The first step is to get the semantic segmentation prototype and then obtain the object detection bounding box.
Afterwards, as its improved version, YOLACT++ \cite{zhou2020yolact++} introduces the deformable convolutional network (DCN) \cite{dai2017deformable, zhu2019deformable} and achieves better results.
Innovatively, TensorMask \cite{chen2019tensormask} uses the structured four-dimensional tensor to represent masks in a dense set of windows, and adopts the mask prediction head and classification head to generate masks and predict categories, respectively.
SOLO \cite{wang2020solo} obtains the corresponding category probability and instance mask on the full convolution feature map directly according to the location and size of instances. After that, SOLOv2 \cite{wang2020solov2} proposes the idea of dynamically segmenting each instance in the image, which divides the task of generating instance masks into mask kernel prediction and mask feature learning.
In addition, CondInst \cite{tian2020conditional}, which is built based on FCOS \cite{Tian_2019_ICCV}, uses instance-aware mask head to predict the instance masks by utilizing conditional convolution, achieving improved performance in both accuracy and inference speed.
Inspired by the class activation maps, SparseInst \cite{cheng2022sparse} proposes the strategy of instance representation by instance activation maps, which greatly reduces the redundant computation.

\subsubsection{Attention-based Instance Segmentation Methods}
Benefiting from its powerful global feature extraction capabilities, attention-based methods have shown excellent performance in the field of instance segmentation.
For instance, ISTR \cite{hu2021istr}, an end-to-end framework for instance segmentation based on the Transformer, adopts the dynamic attention to learn the relations between targets and conducts detection and segmentation concurrently with a recurrent refinement strategy.
Whereas SOTR \cite{guo2021sotr} combines the advantages of both CNN and Transformer to efficiently construct local connectivity and long-range dependencies, which is beneficial for medium and large targets.
In addition, Swin Transformer \cite{liu2021swin, liu2022swin} builds a hierarchical Transformer with shifted windows, which helps the model to extract better feature representation for various vision tasks, including instance segmentation.
To further improve the feature representation ability, MViTv2 \cite{li2022mvitv2} combines the decomposed relative positional embeddings and residual pooling connections with the Vision Transformer (ViT) \cite{dosovitskiy2020image, fan2021multiscale}, which presents excellent performance for object detection and instance segmentation tasks.
Furthermore, Mask DINO \cite{li2023mask} proposes a unified Transformer-based framework for both detection and segmentation, which adds a mask prediction branch that parallels with the detection branch based on the DINO \cite{zhang2022dino}, showing excellent results for instance segmentation on the COCO dataset.

Although these existing methods have achieved promising performance for instance segmentation task in natural images, these methods rarely consider the characteristics of targets in satellite images, making most of them usually encounter difficulties when directly applied to instance segmentation tasks in satellite images.

\section{The Proposed SISP Dataset}
\label{sec:dataset}

\begin{figure*}[htbp]
\centering														
\includegraphics[width=1.0\linewidth]{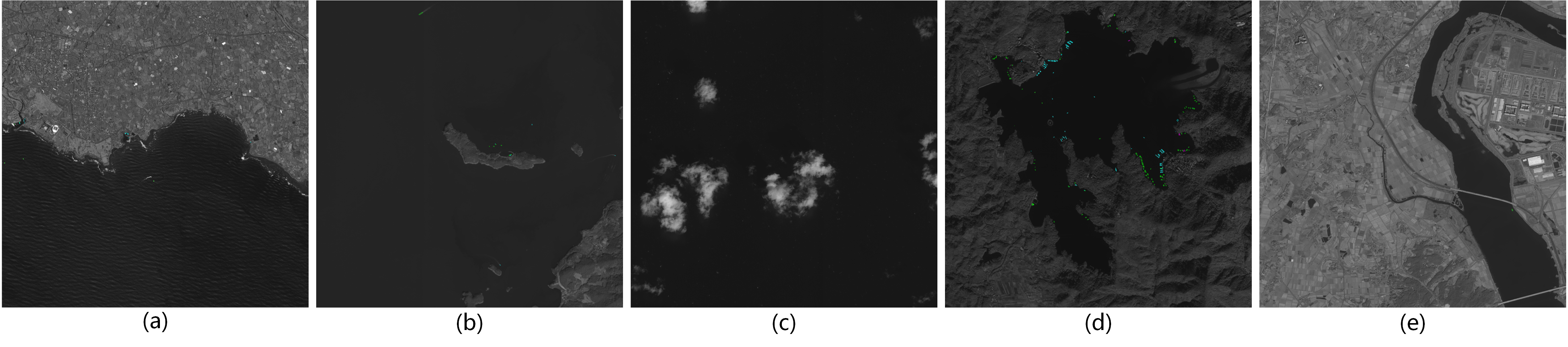}
\caption{Visualization of the collected panchromatic satellite images for the SISP dataset, where (a), (b), (c), (d) and (e) cover different scenarios of the coastal city, island, offshore, lake and river, respectively.}	
\label{fig:img_collection}
\end{figure*}

\subsection{Image Collection and Pre-processing}
\label{subsec:dataset_image_collect}

Considering the requirement of high spatial resolution for the task of ship instance segmentation under satellite observation conditions, the proposed SISP dataset is composed entirely of the high-resolution panchromatic satellite images, where all the images are collected from SuperView-1 satellite, which are panchromatic satellite images with the spatial resolution of 0.5m and the single scene maximum of 60 km $\times$ 70 km.

Specifically, to meet the broad demand of multiple applications as well as to increase the diversity of data, we collect images taken in a variety of scenarios, including coastal cities, islands, offshore, lakes and rivers, as shown in Fig. \ref{fig:img_collection}. In addition, to eliminate the effects of noise in the imaging process of satellite sensors, the imaging quality of the raw data is checked and the images with noise are removed. In this way, we ultimately collect numerous high-quality panchromatic satellite images, in which the size of images ranges from $25,500\times24,900$ to $36,070\times22,870$ pixels.

In addition, due to the large size of panchromatic satellite images and to satisfy the demand of scaling transformation in CNN based methods, all the images are cropped to $800\times800$ pixels with a stride of $550$ pixels. After that, we checked the whole cropped images and filtered out images with pure background. Thus, we can obtain high-quality and high-resolution satellite images.

\subsection{Category Design}

Most existing instance segmentation datasets in satellite images contain only one type for each category, such as building \cite{weir2019spacenet} or ship \cite{wei2020hrsid}. However, targets in satellite images usually belong to multiple fine-grained categories, which makes target recognition more difficult but is more representative of the demands in realistic satellite scenarios. 

Therefore, the selected targets in the SISP dataset include four categories, i.e., civil ship, platform, auxiliary ship and military ship. In the annotation process, the categories of the targets are defined according to their practical functions. Specifically, we define the fishing boats, motorboats, cargo ships, etc., as civil ship, and the marine police ships, crane ships, etc., as auxiliary ship, whereas ships with flat surfaces while without obvious power units are defined as platform. As for military ships, they are judged primarily on characteristics such as launchers as well as aprons. 

\subsection{Image Annotation}
\label{sec:image_annotation}

\subsubsection{Annotation Procedure}

To obtain high-quality annotations for ship instance segmentation in satellite images, all the images of the proposed SISP dataset are annotated by the professional annotators. At the beginning of the annotation process, all the annotators are trained through multiple sessions to clarify the definition of the target categories. Afterwards, all the images are divided into several groups to be annotated by multiple annotators, and all the annotators are asked to annotated all the targets belonging to the selected categories unless the target is too small or the boundaries are too difficult to determine, which can further ensure the efficiency and quality of annotations. 

In addition, the auxiliary information such as geographic coordinates and the corresponding aerial images are also utilized to improve the quality and accuracy of annotations. After the first round of annotating, each group of images is cross-checked by annotators from other groups, where all the possible incorrect annotations will be checked and corrected by the original annotators. Then, experts in the field of satellite imagery interpretation are invited to check all annotations again. After passing all the checks, we can obtain the high-quality annotations.

\subsubsection{Annotation Format}

\begin{figure}[t]
\centering														
\includegraphics[width=0.9\linewidth]{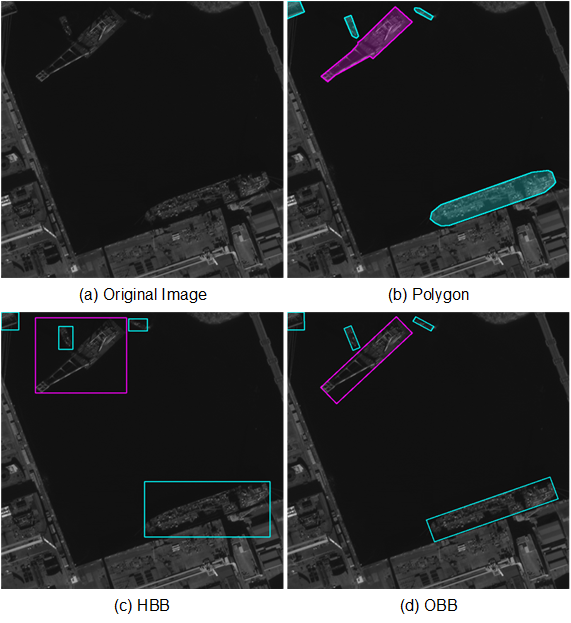}
\caption{Illustration of typical examples from the SISP dataset annotated in different formats, i.e., polygon, HBB and OBB.}
\label{fig:obbhbb}
\end{figure}

The annotation files are produced in the most commonly used annotation format, i.e., COCO format \cite{lin2014microsoft}, where each instance is represented by the polygon, i.e., $[x_1,y_1,x_2,y_2,...,x_n,y_n]$. Here, $(x_i,y_i)$ indicates the location of $i$-th point, and $n$ is the total number of points of the polygon, where $n>2$. In addition, besides the polygon annotation for instance segmentation, our SISP dataset also provides the HBB and OBB annotations for ship detection, as shown in Fig. \ref{fig:obbhbb}. The description of HBB is $(x_c,y_c,w,h)$, where $(x_c,y_c)$ is the central location of each target, and $w$, $h$ are the width and height of the HBB, respectively. Whereas each OBB is described as $(x_c,y_c,w_r,h_r,\theta)$, where $w_r$, $h_r$ are the width and height of the OBB, and $\theta$ indicates the clockwise angle between $w_r$ and $x$-axis. 

\subsection{Statistical Analysis}
\label{sec:data_characteristic}

\begin{figure*}[htbp]
\centering														
\includegraphics[width=1.0\linewidth]{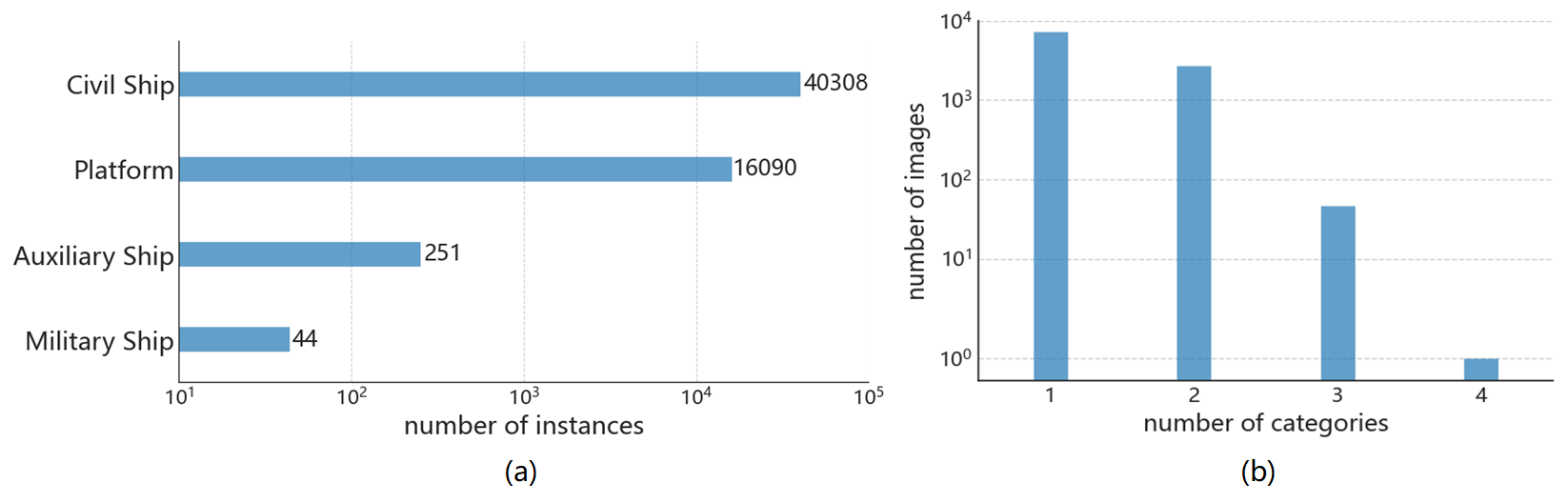} 
\caption{Statistics of category distribution in the SISP dataset. (a) Histogram of the number of instances per category. (b) Histogram of the number of categories per image.}
\label{fig:3_analysis_1}
\end{figure*}

\begin{figure*}[htbp]
\centering
\includegraphics[width=0.9\linewidth]{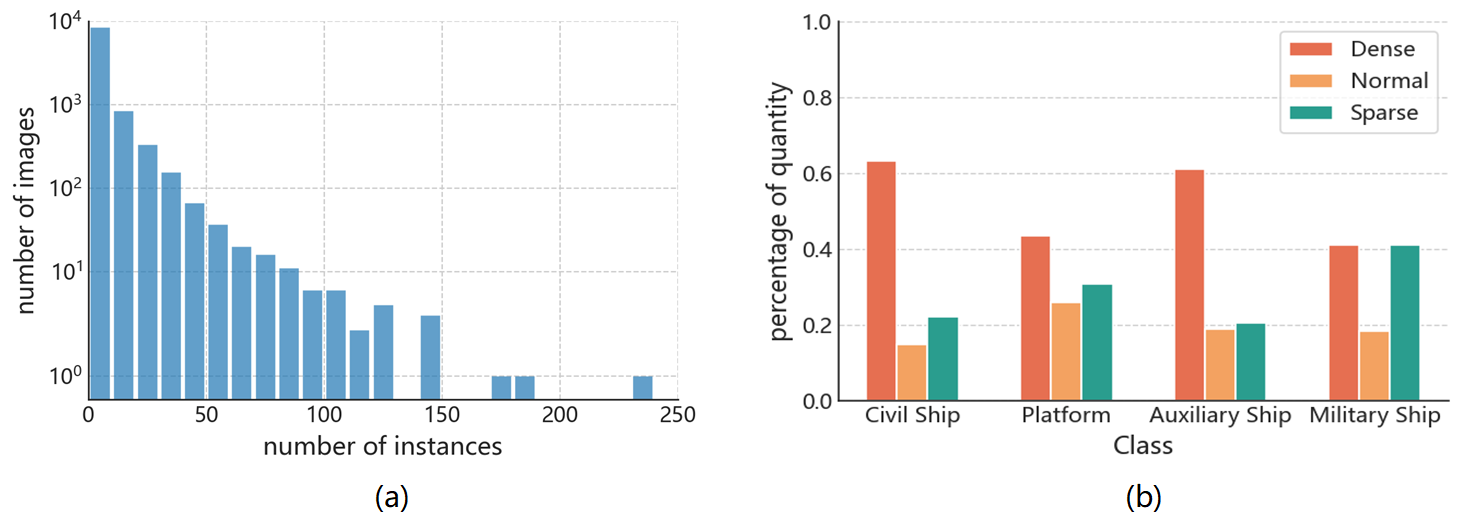}
\caption{Statistics of instances and densities in the SISP dataset. (a) Histogram of the number of instances per image. (b) Histogram of the density distribution for each category. Here, density is calculated by the closet distance between two targets (i.e., polygons), where dense means the distance is less 10 pixels, normal indicates the distance is between 10 and 50 pixels, and sparse means the distance is more than 50 pixels.}
\label{fig:3_analysis_2}
\end{figure*}

\begin{figure*}[htbp]
\centering
\includegraphics[width=0.9\linewidth]{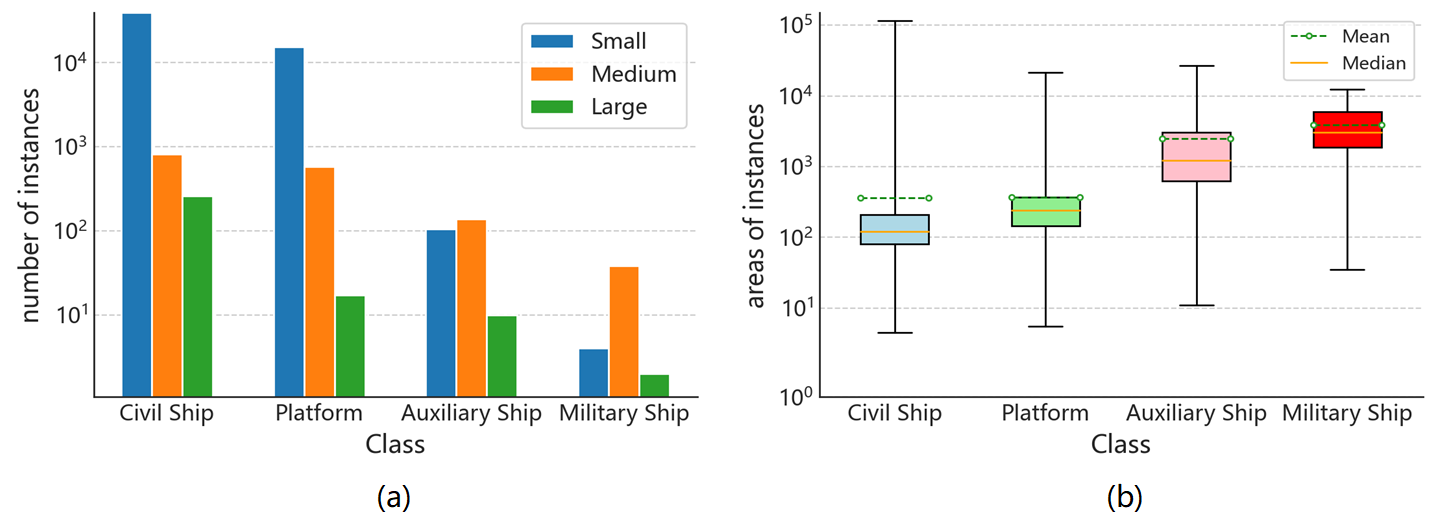}
\caption{Statistics of target size in the SISP dataset. (a) Histogram of the number of instances with different target size for each category, where the area of small target is less than $32^2$ pixels, the area of medium target is between $32^2$ and $96^2$ pixels, and the area of large target is greater than $96^2$ pixels. (b) Boxplot depicting the range of mask areas for each category.}
\label{fig:3_analysis_3}
\end{figure*}

\begin{figure*}[htbp]
\centering														
\includegraphics[width=0.9\linewidth]{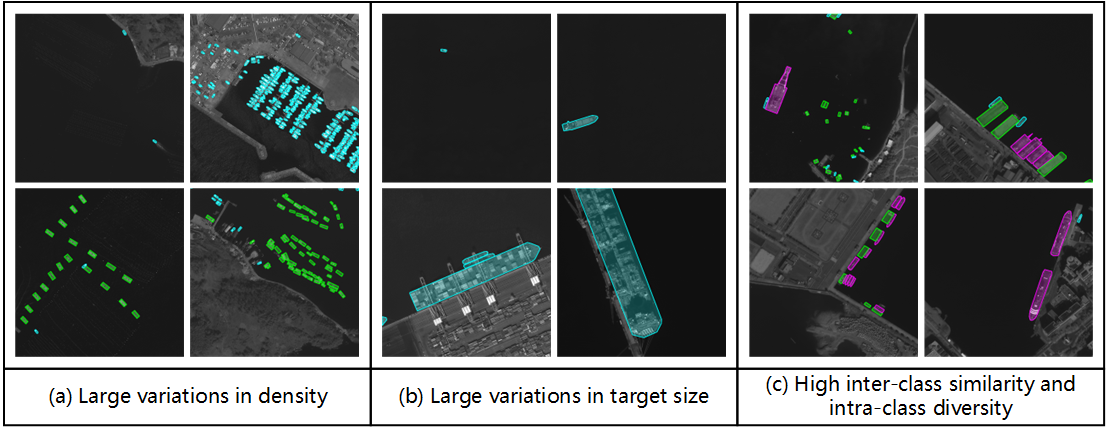}
\caption{Illustration of the characteristics in the SISP dataset, where targets in cyan, green and pink are civil ships, platforms and auxiliary ships, respectively.}
\label{fig:4_characteristic}
\end{figure*}

Although current studies have made significant contributions to release various ship observation datasets in satellite images \cite{wei2020hrsid, airbus-ship-detection, zhang2021shiprsimagenet}, most of them either do not provide pixel-wise mask annotations or fine-grained information, which limits the development of deep learning based ship instance segmentation algorithms for practical satellite applications. The proposed SISP dataset consists of 10,000 sliced panchromatic satellite images and 56,693 well-annotated ship instances. To the best of our knowledge, it is the first fine-grained ship instance segmentation dataset composed entirely of panchromatic satellite images, which is more suitable for real-word application in complicated satellite scenes. 

More statistics and analysis are illustrated in Figs. \ref{fig:3_analysis_1}, \ref{fig:3_analysis_2}, \ref{fig:3_analysis_3} and \ref{fig:4_characteristic}.  The diversity and challenges of the SISP dataset can be reflected in the following characteristics.

\textit{(1) Fine-grained ship categories.} To promote the development of fine-grained ship instance segmentation in satellite images, the SISP dataset provides four fine-grained ship categories, i.e., civil ship, platform, auxiliary ship and military ship, as shown in Fig. \ref{fig:3_analysis_1} (a), whereas existing ship instance segmentation datasets only identify whether a target belongs to the ship. 

\textit{(2) Imbalanced number of instances per category.} As illustrated in Fig. \ref{fig:3_analysis_1} (a), the SISP dataset has a significant long-tail distribution of categories, which contains 40,308 civil ships, 16,090 platforms, 251 auxiliary ships and 44 military ships. Although it reflects the real category distribution in satellite observation, this distribution introduces a serious challenge in identifying categories with few instances, e.g., military ship. In addition, there are usually only one or two categories in a single image, and rarely all categories are present, as illustrated in Fig. \ref{fig:3_analysis_1} (b), due to some categories, such as platform and military ship, rarely appear in the same scene. The challenging category distribution of SISP dataset makes it a better dataset for research on the construction of correlations between different types of ships in real satellite scenarios.

\textit{(3) Large variations in instance densities.} The number of target instances varies significantly in the SISP dataset, as illustrated in Fig. \ref{fig:3_analysis_2} (a). Although most images contain up to 10 instances in a single image, there are still some images that contain more than 100 instances (up to 240 instances per image) in a single image, which makes the instance segmentation task more challenging. We compare images with different instance densities in Fig. \ref{fig:4_characteristic} (a). In addition, to give a quantitative analysis of the density distributions for different categories, we measure the distance of each target from its nearest instance. After that, we divide the targets into three groups based on the distances, i.e., those with distances in $[0, 10)$, $[10, 50)$ and $[50, \infty)$ pixels are dense, normal and sparse targets, respectively. As shown in Fig. \ref{fig:3_analysis_2} (b), most of the targets among different categories in the SISP dataset are densely arranged.

\textit{(4) Large variations in target size.} Because of the huge scale variation of targets themselves, e.g., cargo ships are typically 100m to 400m in size, while motorboats are typically 3m to 4m.  The size of targets in satellite images often varies widely even at the same spatial resolution, as illustrated in Fig. \ref{fig:4_characteristic} (b). Referring to the definition of the target size from COCO metrics \cite{lin2014microsoft}, we consider targets with mask area in the range $[0,32^2)$ as small, $[32^2,96^2)$ as medium and $[96^2,\infty)$ as large. Here, most civil ships and platforms are small targets, while most auxiliary ships and military ships are in medium size, as shown in Fig. \ref{fig:3_analysis_3} (a). In addition, there are a large range of size variations of instances per category in the SISP dataset, as shown in Fig. \ref{fig:3_analysis_3} (b). These large variations in target size increase the complexity of the instance segmentation task, making it more challenging and more suitable for the practical applications.

\textit{(5) High inter-class similarity and intra-class diversity.} Due to the categories in our SISP dataset all belong to ships and are defined at a fine-grained level according to their practical functions, there is inherent high inter-class similarity and intra-class diversity. For instance, as shown in Fig. \ref{fig:4_characteristic} (c), auxiliary ships present different appearances, whereas part of auxiliary ships, e.g., crane ship in folded state, have a very similar appearance to the platforms. The unique characteristics make the SISP dataset suitable for research on fine-grained feature extraction and instance segmentation in satellite images.

\subsection{Dataset Splits}

To facilitate experimental evaluation and comparison, we separate the SISP dataset into train, validation and test splits. Specifically, we randomly select 60\% of the original images as the training set, 20\% as the validation set and the rest as the test set. The distribution of instances in the training, validation and test set for each category in the SISP dataset is listed in Table \ref{tab:split_num}.

\begin{table}[htbp]
    \caption{The distribution of instances in the training, validation and test set for each category in the SISP dataset.}
    \centering
    \begin{tabular}{ccccc}
    \toprule
    Categories      & Train   & Validation   & Trainval  & Test    \\
    \midrule
    Civil Ship      & 24,192  & 8,217        & 32,409    & 7,899   \\
    Platform        & 9,737   & 3,223        & 12,960    & 3,130   \\
    Auxiliary Ship  & 137     & 67           & 204       & 47      \\
    Military Ship   & 29      & 5            & 34        & 10      \\
    \midrule
    Total           & 34,095  & 11,512       & 45,607    & 11,086  \\
    \bottomrule
    \end{tabular}
    \label{tab:split_num}
\end{table}

\section{The Proposed DFRInst Method}
\label{sec:ourmethod}

In this section, we first give the overview introduction of the proposed benchmark method, i.e., dynamic feature refinement-assist instance segmentation network (DFRInst). Then, the three main innovative improvements, i.e., the dynamic feature refinement (DFR) module, the DFR-assist feature pyramid network (DFR-FPN) and the DFR-assist mask head (DFR-MH), are presented in detail.

\begin{figure*}[t]
\centering														
\includegraphics[width=1.0\linewidth]{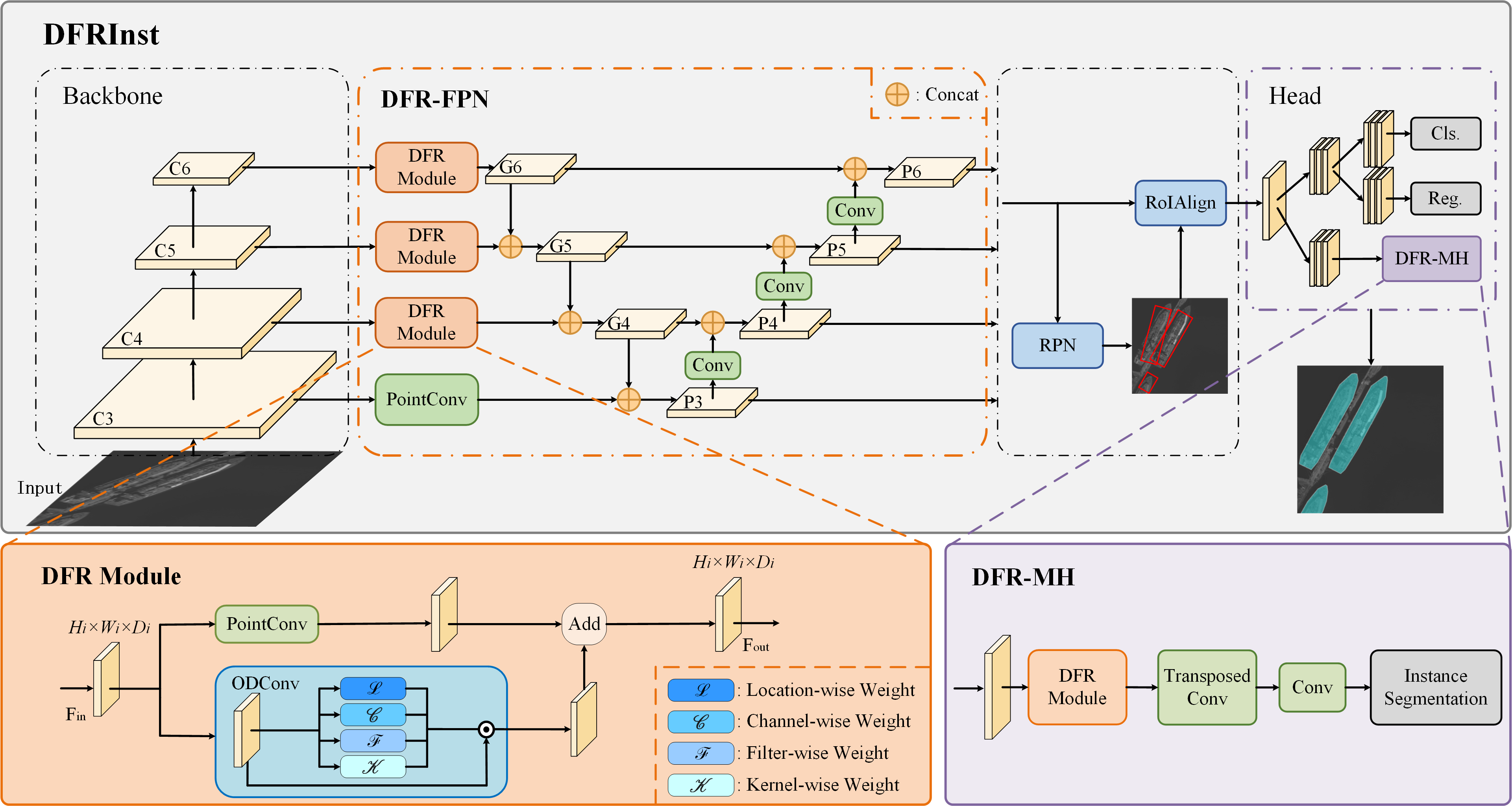}
\caption{
Overview of the proposed DFRInst method. The upper gray block shows the overall network architecture. Here, the DFR module is employed to fortify the explicit representation of crucial features in the image information transmission, which can dynamically explores information from the feature maps with four weight matrices. $\odot$ denotes the multiplication operation along different dimensions of features. In addition, DFR-FPN and DFR-MH are enhanced by the DFR module to improve the perception of salient feature information during feature fusion and decoupling process.}
\label{fig:ourmethod}
\end{figure*}

\subsection{Network Architecture}
We select Mask R-CNN \cite{he2017mask} as our baseline model, which is a classic two-stage instance segmentation framework. Besides, considering the powerful global feature extraction capability of self-attention mechanism, Swin Transformer \cite{liu2021swin} is used as the backbone network.

As illustrated in Fig. \ref{fig:ourmethod}, the upper gray block shows the overall network architecture, which consists of the backbone network, DFR-FPN, region proposal network (RPN), region of interest (RoI) Align and prediction heads. Specifically, let $F_{i} \in \mathbb{R}^{H_{i}\times W_{i}\times D_{i}}$ be the feature maps from the backbone network, where $i$ is the layer index, $H_{i}\times W_{i}$ represents the size of feature maps and $D_{i}$ indicates the dimension of channels. 
Then, following FPN \cite{lin2017feature}, we detect and segment targets with different sizes on different levels of feature maps. Therefore, in this work, DFR-FPN generates four levels of multi-scale feature maps, i.e., $\{P_{3},P_{4},P_{5},P_{6}\}$, from the selected backbone feature maps, denoted as $\{C_{3},C_{4},C_{5},C_{6}\}$.
After obtaining the multi-scale feature maps, following Mask R-CNN, RPN is used to predict candidate regions and RoIAlign is employed to align regional features. Finally, three feature decoupling networks, i.e., classification head, regression head and mask head, are adopted to predict the categories, bounding box regressions and segmentation masks, respectively. 

The above represents the pipeline of the DFRInst, where the three main components of the proposed method, i.e., DFR module, DFR-FPN and DFR-MH, are introduced in detail in the following subsections.

\subsection{Dynamic Feature Refinement Module}
To fortify the explicit representation of crucial features in the image information transmission, we propose the DFR module, which consists of PointConv and ODConv \cite{li2022omni}, as shown in the orange block in Fig. \ref{fig:ourmethod}. Here, PointConv indicates $1\times1$ convolution layer, which is used as a channel compensation strategy to help ODConv in multidimensional information exploration to discover salient features and fuse them into the feature maps. 
ODConv can dynamically explore the information from the feature map to characterize potential targets with four weight matrices, i.e., the location-wise weight $\mathscr{L}$, channel-wise weight $\mathscr{C}$, filter-wise weight $\mathscr{F}$ and kernel-wise weight $\mathscr{K}$. 

In this way, the proposed DFR module achieves the dynamic extension of the feature maps, thus enhancing the target feature representation for more accurate instance segmentation.

\subsection{DFR-assist Feature Pyramid Network} 

To alleviate scale variations, current deep learning based image interpretation methods commonly adopt feature pyramid networks to enhance the fusion of information from different level of feature maps \cite{lin2017feature, liu2018path, lyu2022rtmdet, 10238752}. Therefore, to further facilitate the ability of multi-scale feature extraction for targets in satellite images and emphasize more efficient feature information in the feature fusion process, the DFR-FPN is proposed. 

Specifically, we use the CSP-PAFPN \cite{lyu2022rtmdet} as the baseline FPN network, which achieves excellent performance in natural image interpretation tasks. 
In order to dynamically enhance the information in multi-scale feature fusion process, we employ the DFR module before the feature map of backbone network is transmitted downwards. 

More specifically, let $C_{j}$ be the input feature maps, where $j$ is the layer index. The DFR module is used as the feature preprocess module to introduce addition feature augmentation path, and generates intermediate feature map $G_{j}$. 
As illustrated in Fig. \ref{fig:ourmethod}, the DFR module receives three levels of feature maps of the backbone network, i.e., $C_{4}$, $C_{5}$ and $C_{6}$, and generates $G_{4}$, $G_{5}$ and $G_{6}$, respectively. After that, a series of convolution and concat operations are performed to generate the multi-scale feature maps, i.e., $P_{3}$, $P_{4}$, $P_{5}$ and $P_{6}$.

In this way, the proposed DFR-FPN constructs the top-down and the down-top feature fusion pathways for feature fusion, and the additional feature augmentation path will facilitate the ability of multi-scale feature extraction for targets in satellite images.

\subsection{DFR-assist Mask Head} 
Considering additional effective information propagation, especially the potential information of targets, can significantly improve the performance of segmentation mask prediction. In this work, we introduce the DFR module into mask prediction head to capture more effective information of targets, such as boundary information, and construct the DFR-MH to boost the instance segmentation performance.

Specifically, following Mask R-CNN, we firstly adopt the fully convolutional network (FCN) to decouple aligned regional features obtained from RoIAlign for mask prediction task. After that, the DFR-MH adopts the DFR module, transposed convolution layer and normal convolution layer to generate segmentation masks. Here, the DFR module is used to capture the details associated with potential targets, thus improving the segmentation accuracy. Transposed convolution layer is employed as the up-sampling method to adjust the size of feature maps. After that, the normal convolution layers are adopted to achieve feature decoupling for instance segmentation. 

Therefore, the proposed DFRInst can achieve the accurate ship instance segmentation in satellite images, with potential target information introduction, which is important for models to locate densely arranged targets under satellite observation.

\section{Benchmark Experiments}
\label{sec:experiments}

In this section, representative state-of-the-art instance segmentation methods and the proposed DFRInst are selected as the benchmark methods evaluated on the SISP dataset. 
The details of the experimental setup, evaluation metrics and experimental results are presented as follows. 

\subsection{Experimental Setup}

\subsubsection{Datasets} In our experiments, we use the train set and validation set from the SISP dataset for training, and the test set is used for inference. In order to fairly and directly demonstrate the segmentation performance of the selected benchmark methods on the proposed SISP dataset, we keep the image size at the original size of 800$\times$800 pixels without any date augmentations during both training and inference. In addition, for convenience of description, we use abbreviations for category names, i.e., CS for civil ship, PL for platform, AS for auxiliary ship and MS for military ship.

\subsubsection{Benchmark Methods} For benchmark selection, we have investigated several popular and classic instance segmentation methods and the state-of-the-art methods. As a result, a total of 14 representative instance segmentation models, i.e., 5 two-stage methods, 5 one-stage methods, 3 attention-based methods and our proposed DFRInst, are selected as the benchmark methods, where all the models are collected from the open source code libraries \cite{wu2019detectron2, chen2019mmdetection, tian2019adelaidet}. Specifically, the selected methods are given as following:
\begin{enumerate}[i)]
\item Five two-stage methods, i.e., Mask R-CNN \cite{he2017mask}, Cascade R-CNN \cite{cai2019cascade}, HTC \cite{chen2019hybrid}, PointRend \cite{kirillov2020pointrend} and CenterMask \cite{lee2020centermask}.
\item Five one-stage methods, i.e., YOLACT \cite{bolya2019yolact}, TensorMask \cite{chen2019tensormask}, SOLOv2 \cite{wang2020solov2}, CondInst \cite{tian2020conditional} and SparseInst \cite{cheng2022sparse}.
\item Four attention-based methods, i.e., Swin Transformer \cite{liu2021swin}, MViTv2 \cite{li2022mvitv2}, Mask DINO \cite{li2023mask} and our proposed DFRInst.
\end{enumerate}

It is worth noting that, in our experiments, PointRend is implemented based on Mask R-CNN, while Swin Transformer and MViTv2 are used as the backbone network based on the Mask R-CNN.

\subsubsection{Implementation Details} We keep all the experimental settings the same as the predefined settings in the code libraries except for the number of categories. In addition to Swin Transformer, MViTv2 and DFRInst, we mainly use ResNet50 \cite{he2016deep} pretrained on the ImageNet \cite{deng2009imagenet} as the backbone network in our experiments. All the experiments are implemented on a computer with four NVIDIA A100 GPUs with 40GB memory.

Furthermore, existing deep learning based instance segmentation methods are usually proposed to process conventional optical images with three visible bands, i.e., red, green and blue, whereas the images in the SISP dataset contains a single band, which fails to be used directly by existing methods. Therefore, in our experiments, we duplicate the single channel into three channels and normalize each channel to meet the data requirements of existing methods. 

\begin{table*}[htp]
    \centering
    \caption{The segmentation performance (\%) of 13 representative benchmarks on the proposed SISP test set. Here, Swin-T and MViTv2-T denote the tiny version of models. The best and second-best results are highlighted in bold and underline, respectively.}
    \resizebox{\linewidth}{!}{
    \begin{tabular}{lcccccccccccc}
        \toprule
        Methods & Backbone & AP$_{CS}$ & AP$_{PL}$ & AP$_{AS}$ & AP$_{MS}$ & mAP$_S$ & mAP$_M$ & mAP$_L$ & mAP$_{50}$ & mAP$_{75}$ & mAP$_{50:95}$ & FPS\\
        \midrule
        \textit{two-stage methods: } \\
        Mask R-CNN \cite{he2017mask}        & ResNet50 & 45.07 & 52.96 & 40.23 & 2.36  & 29.46 & 47.23 & 84.93 & 54.72 & 41.43 & 35.15 & 42.2\\
        Cascade R-CNN \cite{cai2019cascade} & ResNet50 & 39.68 & 53.71 & 48.89 & 25.84 & 31.15 & 55.84 & 89.47 & 62.30 & 45.80 & 42.03 & 26.2\\
        HTC \cite{chen2019hybrid}           & ResNet50 & 42.21 & 57.25 & 52.63 & 35.98 & 32.98 & 62.70 & 91.47 & 64.90 & 55.51 & 47.02 & 29.2 \\
        PointRend \cite{kirillov2020pointrend} & ResNet50 & \underline{55.70} & \underline{62.20} & \textbf{65.85} & 32.20 & \textbf{43.14} & 65.61 & \textbf{97.07} & 71.43 & 60.20 & 53.99 & 32.7\\
        CenterMask \cite{lee2020centermask} & ResNet50 & 38.52 & 52.39 & 55.35 & 29.21 & 33.28 & 59.45 & 85.89 & 64.99 & 46.78 & 43.87 & 33.9\\
        \midrule
        \textit{one-stage methods: } \\
        YOLACT \cite{bolya2019yolact}       & ResNet50 & 15.95 & 35.29 & 18.18 &  0.36 & 16.26 & 28.49 & 53.71 & 34.83 & 14.55 & 17.45 & \underline{57.5} \\
        TensorMask \cite{chen2019tensormask} & ResNet50 & 21.65 & 37.31 & 46.14 & 12.72 & 20.73 & 49.37 & 82.57 & 58.38 & 24.49 & 29.46 & 28.9\\
        SOLOv2 \cite{wang2020solov2}        & ResNet50 & 11.11 & 27.28 & 35.58 & 15.99 & 14.18 & 40.94 & 84.92 & 38.66 & 25.15 & 22.49 & 37.7\\
        CondInst \cite{tian2020conditional} & ResNet50 & 21.59 & 40.34 & 48.48 & 22.27 & 22.20 & 55.01 & 90.02 & 63.95 & 30.76 & 33.17 & 33.3\\
        SparseInst \cite{cheng2022sparse}   & ResNet50 & 18.54 & 27.35 & 36.41 & 33.24 & 16.59 & 41.85 & 56.79 & 51.07 & 31.13 & 28.89 & \textbf{66.1}\\
        \midrule
        \textit{attention-based methods: } \\
        MViTv2 \cite{li2022mvitv2}  & MViTv2-T & \textbf{57.72} & \textbf{64.49} & 60.20 & 49.24 & 38.88 & \textbf{72.50} & 91.32 & \textbf{81.41} & \textbf{70.79} & \textbf{57.91} & 24.1\\
        Mask DINO \cite{li2023mask} & ResNet50 & 49.59 & 57.44 & \underline{63.90} & 44.22 & \underline{39.99} & 66.80 & 89.18 & \underline{74.43} & 62.80 & 53.79 & 9.6\\
        Swin Transformer \cite{liu2021swin} & Swin-T & 40.86 & 52.82 & 51.77 & 47.43 & 33.64 & 64.63 & \underline{92.97} & 67.00 & 59.82 & 48.22 & 29.2\\
        \midrule
        \textbf{DFRInst (w/o -F)} & Swin-T & 43.58 & 56.59 & 53.93 & \underline{52.54} & 32.95 & 69.38 & 91.20 & 67.51 & 62.60 & 51.66 & 22.6 \\
        \textbf{DFRInst (w/o -M)} & Swin-T & 43.74 & 56.48 & 58.85 & 49.81 & 36.70 & 68.10 & 90.80 & 73.00 & 59.90 & 52.20 & 21.6 \\
        \textbf{DFRInst (Ours)}  & Swin-T & 43.39 & 57.46 & 58.43 & \textbf{58.29} & 35.80 & \underline{71.40} & 91.00 & 72.80 & \underline{68.40} & \underline{54.40} & 20.5\\
        \bottomrule
    \end{tabular}}
    \label{tab:benchmark}
\end{table*}

\begin{figure*}[htbp]
    \centering
    \includegraphics[width=1.0\linewidth]{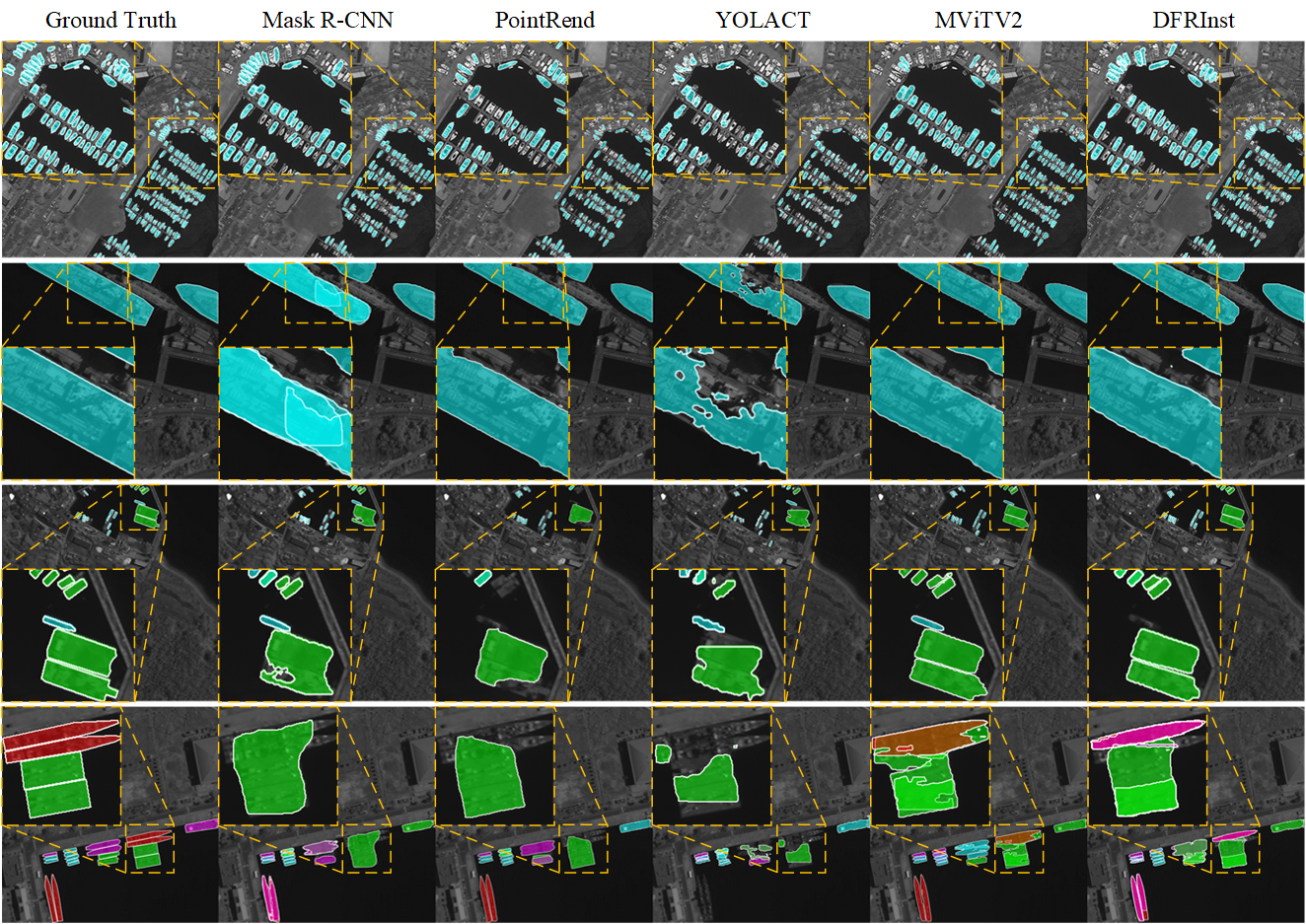}
    \caption{Visualizations of instance segmentation results on the SISP dataset. We have selected typical representatives of the above experimental methods in order to analyze the shortcomings of the current methods on the SISP dataset.}
    \label{fig:5_result}
\end{figure*}

\subsection{Evaluation Metrics}

In order to comprehensively evaluate the performance of the selected instance segmentation benchmark methods, average precision (AP) and mean average precision (mAP) are selected as our evaluation metrics for a single category and the overall performance, respectively. Among them, mAP$_{50}$ and mAP$_{75}$ are adopted to represent the overall performance for the conditions where the intersection over union (IoU) threshold is set as 0.50 and 0.75, respectively. In addition, we use mAP$_{50:95}$ to denote the mean of overall performance with IoU taking values from 0.50 to 0.95 and at intervals of 0.05, i.e., averaged over 10 IoU thresholds. Here, IoU denotes the overlap ratio of predicted mask $D$ and the ground truth mask $G$, which can be calculated as:
\begin{equation}
  \label{eq:iou}
  IoU = \frac{area(G \cap D)}{area(G \cup D)}
\end{equation}
If the IoU is higher than the predefined threshold, the predicted mask is regarded as true positive (TP). Otherwise, it is considered as a false positive (FP). If there is a ground truth mask that does not match any TP, it is denoted as false negative (FN). In this way, we can obtain the number of TP, FP, and FN for the corresponding IoU threshold conditions, and we can calculate the recall and precision as follows:
\begin{equation}
  \label{eq:prcision}
  Precision = \frac{TP}{TP + FP}
  \end{equation}
\begin{equation}
  \label{eq:recall}
  Recall = \frac{TP}{TP + FN}
\end{equation}
After obtaining the precision and recall, the AP and mAP can be calculated based on the COCO \cite{lin2014microsoft} evaluation method, where AP is the precision averaged across different recall levels, while mAP is the mean of AP across all $C$ categories.
\begin{equation}
  \label{eq:mAP}
  mAP = \frac{\sum_{i=1}^{C}AP_i}{C}
\end{equation}

Furthermore, for better understanding the segmentation performance for targets with different sizes, mAP$_S$, mAP$_M$ and mAP$_L$ are used to represent the performance for small, medium and large targets, with range of $[0,32^2)$, $[32^2,96^2)$ and $[96^2,\infty)$ pixels, respectively.
The frames per second (FPS) is used to evaluate the inference speed.

\subsection{Experimental Results and Analysis} 

The benchmark results on the SISP dataset are given in Table \ref{tab:benchmark}. Here, the components of proposed DFRInst are indicated in abbreviated form, i.e., `-F' indicates the DFR-FPN and `-M' refers to the DFR-MH. `w/o' is applied to indicate without specified module. The visualization of instance segmentation results for several typical methods on the test set is illustrated in Fig. \ref{fig:5_result}. 

Firstly, the performance of benchmarks on all the categories verifies the difficulty of our dataset. Different from generic categories, the fine-grained ship categories are characterized by high inter-class similarity and intra-class variability, leading to poor performance in most methods.
Besides, class imbalances have great impacts on the ship segmentation performance. Specifically, due to the few number of military ships, most instance segmentation methods can hardly extract its rich feature representations, resulting in low accuracy compared with other categories with sufficient number of instances. For instance, Mask R-CNN achieves 45.07\% AP$_{CS}$ but only 2.36\% AP$_{MS}$. 
Whereas our DFRInst achieves the best performance of 58.29\% AP$_{MS}$, which demonstrates that the boundary information facilitates fine-grained ship segmentation to some extent.

In addition, it can be observed that all the benchmarks have poor segmentation performance for small targets. As shown in Table \ref{tab:benchmark}, the highest AP for small targets is only 43.14\%, while the highest AP for large targets is 97.07\%. The results illustrate that existing instance segmentation methods, especially the one-stage approaches, are often difficult to accurately localize small targets due to the difficulty of extracting discriminative features of small targets. Therefore, the problem of small targets not being segmented is more noticeable, as shown in the first row of Fig. \ref{fig:5_result}. However, the proposed SISP dataset contains a large number of small targets posing a great challenge to existing instance segmentation algorithms. 
As for different types of methods, it can be observed that two-stage algorithms typically achieve higher segmentation accuracy compared with one-stage algorithms, as show in the second row of Fig. \ref{fig:5_result}. Whereas attention-based methods usually achieve better performance benefiting from the robust global feature extraction.

Specifically, compared with other two-stage methods, PointRend achieves the best segmentation performance of 71.43\% mAP$_{50}$, 60.20\% mAP$_{75}$ and 53.99 mAP$_{50:95}$. This results demonstrate that PointRend can predict more accurate pixel-wise masks with an iterative rendering strategy. As for one-stage methods, CondInst achieves promising results, but there is still a large gap in accuracy compared with the other two types of instance segmentation methods. As for attention based methods, the MViTv2 achieves the best performance among all benchmark methods of 81.41\% mAP$_{50}$, 70.79\% mAP$_{75}$ and 57.91\% mAP$_{50:95}$, which demonstrates that extracting more robust and stronger features can help the model to achieve more accurate instance segmentation results. 

Besides, compared with Swin Transformer, which is the baseline of our proposed method, DFRInst (w/o -F) and DFRInst (w/o -M) have improved significantly in almost every metrics, which further demonstrate the effectiveness of the ODFPN and ODMaskHead. In addition, as shown in the third and last rows of Fig. \ref{fig:5_result}, most methods are often difficult to segment densely arranged targets, leading to the prediction of two side-by-side targets as a single instance and resulting in decreased accuracy. Whereas our DFRInst can predict the pixel-wise masks for the densely arranged targets more accurately, which obtains a more robust feature representation by capturing and fusing boundary information. The results prove the effectiveness of the proposed DFRInst.

In general, the above benchmark results demonstrate that the SISP dataset presents numerous challenges due to its characteristics in terms of instance segmentation task in satellite images. Nevertheless, these challenges are more reflective of the complexities of real-world satellite scenarios. Therefore, the proposed SISP dataset can facilitate the study of fine-grained ship instance segmentation in satellite images.

\section{Conclusion}
\label{sec:conclusion}
In this paper, we proposed a publicly available dataset for fine-grained ship instance segmentation in panchromatic satellite images, namely SISP, and introduced a benchmark method for this dataset. The proposed dataset has the unique characteristics, such as high class imbalances, various scenes, large variations in target densities and scales, and high inter-class similarity and intra-class diversity, which make it more suitable for real-world applications in complicated satellite scenes. The process of image collection and annotation, as well as the analysis of statistics are presented in detail. By taking into account the characteristics of targets in the SISP dataset, our proposed benchmark method can fortify the explicit representation of crucial features, thus improving the performance of ship instance segmentation in satellite images. In addition, several representative benchmark methods are evaluated on the SISP dataset, which can help researchers to build the baselines for fine-grained ship instance segmentation under satellite observation. Our proposed SISP dataset and the benchmark evaluations will benefit the development of the fine-grained ship instance segmentation in satellite images.

\bibliographystyle{IEEEtran}
\bibliography{cas-refs}

\vfill

\end{document}